\documentclass{article}

\usepackage[final, nonatbib]{neurips_2021}

\usepackage[utf8]{inputenc} 
\usepackage[T1]{fontenc}    
\usepackage{hyperref}       
\usepackage{url}            
\usepackage{booktabs}       
\usepackage{amsfonts}       
\usepackage{nicefrac}       
\usepackage{microtype}      
\usepackage{xcolor}         
\usepackage{graphicx}
\usepackage[sorting=none]{biblatex}
\usepackage{todonotes}
\usepackage{comment}
\usepackage{soul}
\usepackage{float}
\usepackage{hyperref}

\title{Multi-lingual agents through multi-headed neural networks}

\author{%
Jonathan D. Thomas \\
University of Bristol\\
\texttt{jt17591@bristol.ac.uk} \\
\And
Ra\'{u}l Santos-Rodr\'{i}guez\\
University of Bristol \\ 
\texttt{enrsr@bristol.ac.uk}\\
\And
Robert Piechocki \\
University of Bristol\\
\texttt{r.j.piechocki@bristol.ac.uk}\\
\And
Mihai Anca \\
University of Bristol\\
\texttt{mihai.anca@bristol.ac.uk}\\
}

\begin{document}

\maketitle

\begin{abstract}
This paper considers cooperative Multi-Agent Reinforcement Learning, focusing on emergent communication in settings where multiple pairs of independent learners interact at varying frequencies. In this context, multiple distinct and incompatible languages can emerge. 
When an agent encounters a speaker of an alternative language, there is a requirement for a period of adaptation before they can efficiently converse. 
This adaptation results in the emergence of a new language and the forgetting of the previous language.
In principle, this is an example of the Catastrophic Forgetting problem which can be mitigated by enabling the agents to learn and maintain multiple languages.
We take inspiration from the Continual Learning literature and equip our agents with multi-headed neural networks which enable our agents to be multi-lingual. 
Our method is empirically validated within a referential MNIST based communication game and is shown to be able to maintain multiple languages where existing approaches cannot.
\end{abstract}

\section{Introduction}

Questions pertaining to communication naturally arise when considering Multi-Agent systems. 
It is natural as communication is such a vital part of our societies, enabling for the dissemination of ideas and large-scale coordination. 
By equipping agents with capacity to communicate they will likely be able to achieve greater levels of synergy with both artificial and biological entities. 

This paper focuses on emergent communication within multi-agent reinforcement learning (MARL), specifically addressing settings where agents can be considered as independent learners (IL)\cite{tan}. 
This restriction removes common methodologies which are utilised to improve training speed and stability, such as centralised training decentralised execution (CDTE) \cite{foersterdialrial}, parameter sharing \cite{tan} and gradient propagation through other agents. 
This is justified by the motivation of creating algorithms that better approximate human learning, where, for example, models of other agents are unlikely to be available for gradient propagation. 
Recent work has attempted to improve training efficiency through a variety of methods. 
\citeauthor{jaques} \cite{jaques} proposes an intrinsic reward based on social influence to encourage communication of useful information and \citeauthor{eccles} \cite{eccles} proposes the introduction of biases to promote the emergence of communication.

In this more natural setting, experimentation has generally been restricted to two independent agents. 
However, realistic scenarios are likely to involve larger numbers of independent agents interacting at varying frequencies. 
As the agents do not use parameter sharing, it is conceivable that multiple unique languages may arise where these languages are unlikely to be compatible.
As result of this, any interaction with a new agent mandates the learning of a shared language.
Without specific modifications to the agent's architecture, this new language will overwrite the previous one as a consequence of a known phenomena within machine learning (ML) named catastrophic forgetting \cite{catforget}. 
As the previous language has been lost, any interaction with the associated conversational partner will require re-training. Here, in order to address this issue, architectural modifications inspired by the Continual Learning literature are introduced into the base algorithm proposed by \citeauthor{eccles} \cite{eccles}.
Namely, multi-headed neural networks are used where a different head is maintained for each language. 
This paper formalises this concept and demonstrates the benefit within a simple MNIST-based referential game (as used within \cite{eccles}).

\section{Related Work}


The general challenge of inter-agent communication has attracted much attention within the MARL community. 
A variety of approaches have been recently proposed.
A few of the most relevant include RIAL \cite{foersterdialrial}, DIAL \cite{foersterdialrial}, CommNET \cite{commnet}, TarMAC \cite{tarmac} or DGN \cite{dgn}.
Works in the area of inter-agent communication can be loosely categorised into two main types, namely those that allow gradients to flow between agents and those that do not.
Recently, there has been interest in the latter domain, whereby facilitation of centralised training and parameter sharing are removed and agents are only allowed to train via the environment reward. 
This is sometimes referred to as Independent Learners \cite{tan}. 

\paragraph{Independent Learners in emergent communication.} 
Despite the additional difficulty, it is often argued that this is a more realistic setting as it is closer to the methods by which humans learn. State-of-the-art examples include \cite{jaques} and \cite{eccles}. 
In \cite{jaques}, an intrinsic reward derived from causal influence is used to encourage the speaker to send messages that change the listeners policy. 
Differently, \cite{eccles} introduces biases into both the speaker and the listener. 
Here, the speaker is encouraged to maximise mutual information between its observation and its message while the listener is encouraged to modify its policy in response to the reception of a message. While both methods are related, following \cite{lowe}, we can summarise \cite{eccles} as encouraging positive signalling and positive listening whereas \cite{jaques} only encourages positive signalling. In this paper we use \cite{eccles} as a baseline for our experimental work, as it can be shown to outperform \cite{jaques} in our setting.

\paragraph{Zero-shot coordination.} A related area of growing interest is zero-shot coordination (ZSC) \cite{hu_zsc,treutlein_zsc,bullard_zsc}, where the objective is to derive policies for cooperative settings which allow for previously unseen partners.
\citeauthor{hu_zsc} \cite{hu_zsc} consider issues posed by the standard self-play methodology where learnt policies are not compatible with novel partners due to agents not being able to exploit potential known symmetries in coordination tasks.
They propose the  \textit{Other-play} algorithm (OP), which involves techniques based on domain randomisation.
\citeauthor{treutlein_zsc} \cite{treutlein_zsc} build upon \cite{hu_zsc}, formalising the setting as a label-free coordination problem (LPCB). 
Finally, \citeauthor{bullard_zsc} \cite{bullard_zsc} explicitly consider communication within ZSC. 
The setting they study involves a costed communication channel with a non-uniform distribution over messaging intents.  Based on OP, they introduce  Quasi-Equivalence Discovery (QED).

Our work elaborates upon previous contributions within the emergent communication literature. 
We follow a deviation from the standard ZSC setting as in \cite{bullard_zsc}. 
In our setting multiple pairs of speakers and listeners are allowed to develop potentially unique languages.
We then address how to both learn and maintain multiple languages and the mitigation of the issues introduced by catastrophic forgetting.

\section{Setting}

The MARL approach defined within this paper is applied to an $N$-player partially-observable Markov game \cite{shapley}, $G$.
Where $G$ is defined by the tuple $G = (S, A_{1}, ..., A_{n}, M_{1}, ..., M_{n} T, O, r)$.
The environment state is defined by $s \in S$. 
At each time-step each agent makes a local observation of the environment state according to $O: S \rightarrow o$. 
In addition to an agent's observation $o$, it also receives all messages from the previous time-step $\mathbf{m}$ (excluding it's own message).
Using this information agents select an action $a_{i} \in A_{i}$ according to $\pi_{i, a}$ and a discrete message $m \in M_{i}$ according to the policy, $\pi_{i, m}$. 
All agents actions make up the joint action $\mathbf{A}$, which results in a state transition according to $T: S, \mathbf{A} \rightarrow S$ and all agents receive a reward $r: S, \mathbf{A} \rightarrow \mathcal{R}$. 
This work is constrained to fully-cooperative games where communication is provably advantageous.
Agents are tasked with finding action policies $\pi_{i,a}: (o, m) \rightarrow A_{i}$ and a message policy $\pi_{i,m}: (o,m) \rightarrow M_{i}$ such that the cumulative discounted reward is maximised.

\section{Method}
\label{sec:methodology}
\subsection{Problem Statement}
\label{sec:methodogy_prob}
Let us consider the existence of two sets of agents, where these are referred to as speakers $T_x =  \{\pi_{s,0}, ... \pi_{s,n}\}$ and listeners $R_x = \{\pi_{l,0}, ... \pi_{l,n}\}$, respectively\footnote{To avoid clashes with standard RL notation, the speakers and listeners have symbols consistent with transmitter and receiver.}.
All agents are parameterized by deep neural networks (DNN) according to the methodology described by \citeauthor{eccles} \cite{eccles}, where this includes introduction of inductive biases to promote the emergence of communication.
For some pairing of $T_x$ to $R_x$, the agents capacity to effectively convey information will be limited by their ability to understand one another. 
Overtime, the agents can adapt to each other and arrive at an emergent protocol which maximises task reward.

The first question this work intends to delve into is, what happens to their established emergent protocol when an agent (be that the speaker or the listener) interacts with a new partner?
More formally, when the mapping from $T_x$ to $R_x$ is randomised and a period of training is allowed, how does this impact the agent's capacity for conversation with its previous partner? 
This problem exists within the continual learning setting, where Catastrophic Forgetting is known to be an issue \cite{catforget}.
It should be expected that as a pair of agents build up familiarity with one another, their previous languages will drift.

The fundamental issue with this mode of operation is that it always requires an agent to re-train upon interacting with a different partner even if they had previously arrived at an efficient protocol. Ideally, this should be avoided as this period of adaptation is costly. 
Naturally, the second question is simply, how can we mitigate this issue? 

\subsection{Multi-headed agents}
\label{sec:algo}

As mentioned above, this primary issue in our scenario is Catastrophic Forgetting \cite{catforget}.
Following the naming convention from \cite{contlearnsurvey}, our approach considers a simple parameter isolation method, where each speaker and listener maintains a separate output head for each possible partner. This idea is based on \cite{decaf}.
It is assumed that the identity of each potential partner is observable and therefore the correct head can be chosen.

\begin{figure}[H]
\centering
\includegraphics[width=10cm]{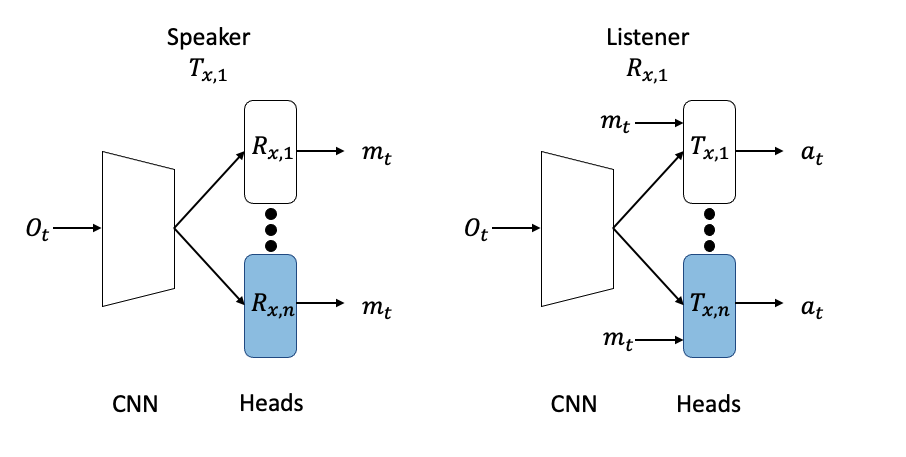}
\caption{DNN architecture of speaker and listener shown on the left and right, respectively.
The networks maintain a separate head for each possible partner, where the label indicates the conversational partner that it refers to.  
The white and blue colouring is representative of how gradients are allowed to propagate through the network. 
In both cases the CNN and first head are trained together, whereas the alternative heads are trained separately.  
}
\label{fig:arch}
\end{figure}

The architecture is presented in Fig.~\ref{fig:arch}, where the CNN for both the speaker and listener are only trained with the first partner. 
This decision is justified by the assumption that, in most cases, languages consider mappings from a similar set of concepts to different words or phrases and, as such, the features learned by the CNN for one language should be transferable. 
An additional variant upon this model is proposed in which the weights of the non-primary heads are pre-initialised with those of the primary head upon establishment of the first language. 
This can be demonstrated to improve sample efficiency when compared to random initialisations.

\section{Experiments}
\label{sec:experi}

\subsection{Implementation}

All code is implemented in Pytorch \cite{pytorch} according to the methodology described in Section \ref{sec:methodology}\footnote{Code available at \url{https://github.com/Jon17591/multi-lingual-agents}}.
As previously introduced, the implementation of the speaker and listener follows the methodology described by \citeauthor{eccles} \cite{eccles}, where we train agents independently utilising REINFORCE and utilise the same hyperparameters.
As we were unable to achieve convergence with the defined architecture we made one modification. 
We introduced an extra layer into the DNN which alleviated this issue.
This minor modification to the method proposed by \citeauthor{eccles} \cite{eccles} without the multi-headed output is utilised as a baseline within our experimentation.

\subsection{Communication Carousel}
\label{sec:game}
\begin{figure}[h]
\centering
\includegraphics[width=12cm]{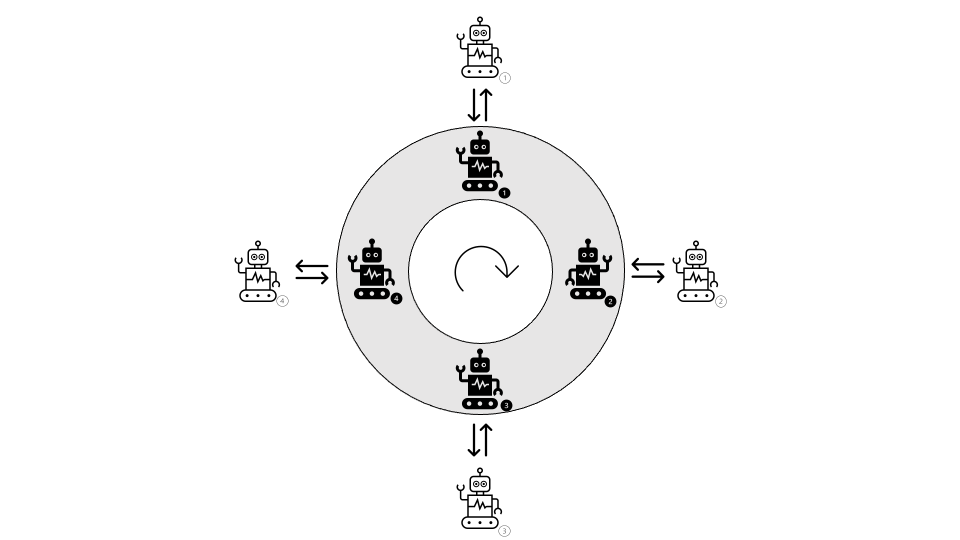}
\caption{Illustration of the $N$-parallel referential games. After $E$ episodes, the carrousel rotates and all agents interact with a different partner. This continues for the desired number of rotations after which all agents are returned to their original partner for assessment of emergent language maintenance.}
\label{fig:agent_rotate}
\end{figure}

The intention of this work is to investigate the agents' capacity to maintain emergent languages after interacting with new partners.
To achieve this, $N$-parallel referential games are instantiated and speakers and listeners are afforded $E$ episodes with their initially assigned partner. 
After the initial $E$ episodes, the agents are rotated and allowed the same number of episodes to interact with their new partner. 
This is illustrated in Figure \ref{fig:agent_rotate}.
After a number of partner changes, $\omega$, the speakers and listeners are returned to their initial partner and afforded a further $E$ episodes to reconverge. 
All experimental parameters are introduced in Table \ref{tab:experi_params}.
This environment formulation provides a simple and interpretable test-bed for studying agent adaptation where the complexity can be easily controlled through appropriate selection of the referential game.

The referential game maintains broadly the same structure as \citeauthor{eccles} \cite{eccles} which is a simple MNIST based game. 
It comprises of two agents, a speaker and a listener who are both presented with images from the MNIST dataset \cite{mnist}. 
The speaker's input is an image from the dataset and it's output is a discrete discrete message $m_t$ which gets passed to the listener.
The listener observes it's own image and the speaker's message and is tasked with adding the two together, where it's answer is represented by it's action $a_t$. 
If the action is equal to the summation of the digits both agents receive a reward of $1$, otherwise the reward is $-1$.
By design, this game can only be successfully completed if an effective language is derived. 

\begin{table}[H]
\caption{Table depicting experimental parameters used in carousel environment}
\label{tab:experi_params}
\centering
\begin{tabular}{lll}
\toprule
Name     & Symbol  & Value \\
\midrule
$N$ & Number parallel environments  & $4$     \\
$E$ & Episodes per interaction & $75$k     \\
$\omega$ & Number of rotations & $1$ \\ 
\bottomrule
\end{tabular}
\end{table}

\section{Results and Discussion}

The results obtained support the hypothesis that the Multi-headed methods defined within Section \ref{sec:algo} results in better maintenance of multiple emergent languages.

\begin{figure}[H]
\centering
\includegraphics[width=10cm]{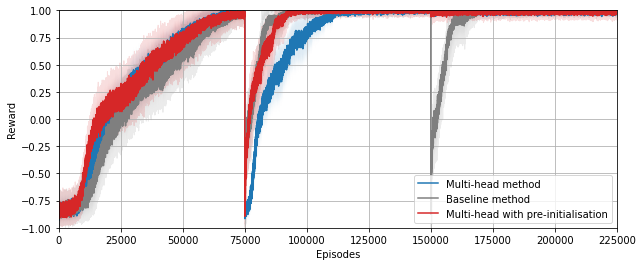}
\caption{Average reward obtained by all 4-agents with their current partner. Partner is changed to a new partner at 75000 episodes and then to the original partner at 150000 episodes. }
\label{fig:training}
\end{figure}

\begin{figure}[H]
\centering
\includegraphics[width=10cm]{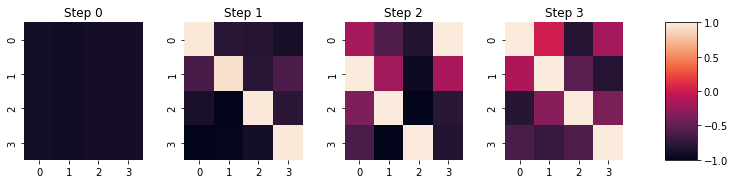}
\caption{Heatmap for baseline method evaluated for all pairings at episodes=0, 75000, 150000 and 225000. Scale represents the average reward which is obtained over 100 episodes.}
\label{fig:baseline_cm} 
\end{figure}

\begin{figure}[H]
\centering
\includegraphics[width=10cm]{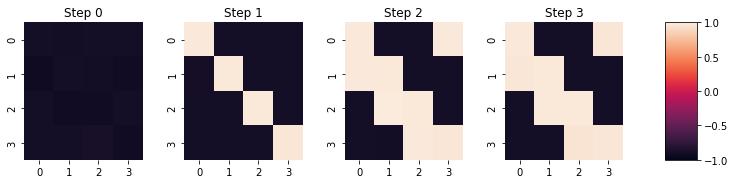}
\caption{Heatmap for Multi-head method evaluated for all pairings at episodes=0, 75k, 150k and 225k. Scale represents the average reward which is obtained over 100 episodes.}
\label{fig:multihead_cm}
\end{figure}

\begin{figure}[H]
\centering
\includegraphics[width=10cm]{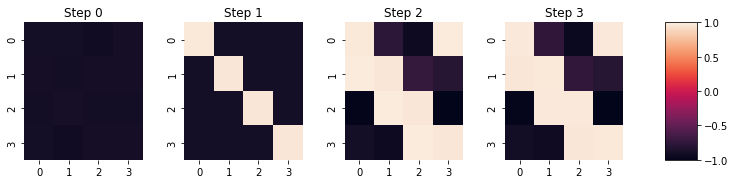}
\caption{Heatmap for Multi-head method with pre-initialisation evaluated for all pairings at episodes=0, 75k, 150k and 225k. Scale represents the average reward which is obtained over 100 episodes.}
\label{fig:multihead_init_cm}
\end{figure}

Figure \ref{fig:training} demonstrates the average reward which agents receive with their current conversational partner for the baseline, Multi-headed method and the Multi-headed method with pre-intialisatation of the non-primary heads.
The most notable observation to draw from this Figure is that the reward for the baseline method reduces substantially when it returns to the initial conversational partner at $150k$ episodes, this reduction is not present in either of the Multi-headed method.
This would suggest that Catastrophic Forgetting has been avoided.
This claim is further supported by Figure \ref{fig:baseline_cm},  \ref{fig:multihead_cm} and \ref{fig:multihead_init_cm}. 
These figures show the average reward obtained by all pairings of speakers and listeners in the form of a heatmap. 
The steps refer to the beginning of training, after every partner switch and at the end of training where this corresponds to episodes $0$, $75$k, $150$k and $225$k in Figure \ref{fig:training}.
Note that the baseline method experiences a significant reduction in reward acquisition once it has trained with a new partner whereas this is not present in either of the Multi-headed methods.

\begin{figure}[h]
\centering
\includegraphics[width=10cm]{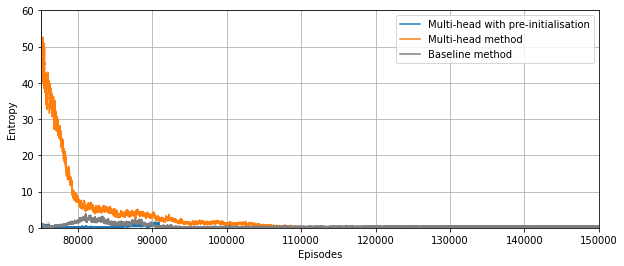}
\caption{Entropy of speakers between $75$k and $150$k episodes.}
\label{fig:entropy}
\end{figure}

A drawback of the standard Multi-headed method appears to be the reduction in sample efficiency present when switching to the second partner ($75$k episodes) in Figure \ref{fig:training}.
It seems that the Multi-headed method takes longer to acquire the second language.
This is as the additional heads are untrained and comprise of randomly initialised weights. 
The baseline method represent a policy that has converged to a solution. 
The entropy of both sets of speaker policies (shown in Figure \ref{fig:entropy}) gives an indication as to why this occurs. 
It is clear that the Multi-headed method begins with significantly higher entropy. 
The introduction of this extra stochasticity may make the arrival at a common protocol more time intensive as there is less determinism to the respective messages and, as such, it is more difficult to achieve synchronisation between the agents.
This can be overcome by pre-initialising the weights of each head with the solution of the primary head, thereby achieving comparable convergence speeds to the baseline.

A further observation is that all the languages are unique. 
The resulting multi-agent system is one with a quadratic relationship between the number of languages and the number of speakers/listeners.
This is not the case in natural systems with the number of distinct languages being somewhat restricted. 
An interesting avenue to further explore involves methodology by which the number of languages can be restricted, aiming to improve zero-shot performance.

\section{Conclusion}
We consider the development of agents which can maintain multiple languages without falling victim to catastrophic forgetting.
This work builds upon that by \citeauthor{eccles} \cite{eccles} and introduces a parameter isolation method into their neural network in order to mitigate the aforementioned issues. 
The modification involves the utilisation of a multi-headed output network, where each head is utilised for a specific language. 
This approach was validated empirically within a novel referential game formulation which facilitated evaluation of language maintenance through interactions with multiple unique agents and will serve as a simple test-bed for future work. 
The results demonstrate that the proposed method effectively avoids catastrophic forgetting when compared to the standard implementation of \citeauthor{eccles} \cite{eccles}. 
Future work intends to consider this methodology within more complex domains and zero-shot scenarios.

\begin{ack}
This work is funded by the Next-Generation Converged Digital Infrastructure (NG-CDI) Project, supported by BT and
Engineering and Physical Sciences Research Council (EPSRC), Grant ref. EP/R004935/1. RSR is partially funded by the UKRI Turing AI Fellowship EP/V024817/1.
\end{ack}

\printbibliography

\end{document}